\documentclass[journal]{IEEEtran}
\usepackage{graphicx} 
\usepackage{subfig}
\usepackage{amsmath}
\usepackage{amsthm}
\usepackage{lipsum}
\usepackage{mathtools}
\usepackage{float}
\usepackage{caption}
\usepackage{color}
\usepackage{placeins}
\usepackage{float}
\usepackage{tabularx,colortbl}
\usepackage{amsmath}
\usepackage{listings}
\usepackage{epstopdf}
\usepackage{amssymb}
\usepackage{algpseudocode}
\usepackage{url}
\usepackage[linesnumbered,ruled]{algorithm2e}
\usepackage{algcompatible}

\usepackage{enumitem}
\usepackage{multicol, blindtext}
\usepackage[american]{babel}
\usepackage{microtype}
\makeatletter

\begin{document}

\title{Vision-based Real-Time Aerial Object Localization and Tracking for UAV Sensing System}

\author{Yuanwei~Wu, \IEEEmembership{Student~Member, ~IEEE,}
		Yao~Sui,~\IEEEmembership{Member,~IEEE,}      
        and~Guanghui~Wang,~\IEEEmembership{Member,~IEEE}
\thanks{Y.~Wu, Y.~Sui, and~G.~Wang are with the Department of Electrical Engineering and Computer Science,
	University of Kansas, 1520 West 15th Street,
	Lawrence, KS 66045. Email: \textsl{ghwang@ku.edu, wuyuanwei2010@gmail.com}.}}

\maketitle

\begin{abstract}
The paper focuses on the problem of vision-based obstacle detection and tracking for unmanned aerial vehicle navigation. A real-time object localization and tracking strategy from monocular image sequences is developed by effectively integrating the object detection and tracking into a dynamic Kalman model. At the detection stage, the object of interest is automatically detected and localized from a saliency map computed via the image background connectivity cue at each frame; at the tracking stage, a Kalman filter is employed to provide a coarse prediction of the object state, which is further refined via a local detector incorporating the saliency map and the temporal information between two consecutive frames. Compared to existing methods, the proposed approach does not require any manual initialization for tracking, runs much faster than the state-of-the-art trackers of its kind, and achieves competitive tracking performance on a large number of image sequences. Extensive experiments demonstrate the effectiveness and superior performance of the proposed approach.
\end{abstract}


\section{Introduction}\label{sec:Introduction}
In the last two decades, we have seen rapid growth in the applications of unmanned aerial vehicles (UAV). In the military, UAVs have been demonstrated to be an effective mobile platform in future combat scenarios. In civil applications, numerous UAV platforms have mushroomed and been applied to surveillance, disaster monitoring and rescue~\cite{fasano2014morpho}, package delivering~\cite{AmazonDrone2013}, and aerial photography~\cite{lyu2016vision}. A number of companies are developing their own UAV systems, such as Amazon Prime Air~\cite{AmazonDrone2013}, Google's Project Wing~\cite{GoogleProjectWing}, and DHL's Parcelcopter~\cite{DHLParcelcopter}. In order to increase the flight safety, the UAV must be able to adequately sense and avoid other aircraft or intruders during its flight. 

The ability of sense and avoid (SAA) enables UAVs to detect the potential collision threat and make necessary avoidance maneuvers. This technique has attracted lots of attention in recent years. 
Among all available approaches, vision-based SAA system~\cite{lyu2016vision,mejias2016sense} is becoming more and more attractive since cameras are light-weighted and low-cost, and most importantly, they can provide richer information than other sensors. A successful SAA system should have the capability to automatically detect and track the obstacles. The study of these problems, as a central theme in computer vision, has been active for the past decades and achieved great progress.  

Salient object detection in computer vision is interpreted as a process of computing a saliency map in a scene that highlights the visual distinct regions and suppresses the background. Most salient object detection methods rely on the assumption about the properties of objects and background. The most widely used assumption is contrast prior~\cite{cheng2015global,jiang2013salient,jiang2013submodular}, which assumes that the appearance contrasts between the objects and backgrounds are very high. Several recent approaches exploit image background connectivity prior~\cite{zhu2014saliency,zhang2015minimum}, which assumes that background regions are usually connected to the image boundary. However, those methods lack of the capability to utilize the contextual information between consecutive frames in the image sequence. 

On the other hand, given the position of the object of interest at the first frame, the goal of visual tracking is to estimate the trajectory of the object in every frame of an image sequence. The tracking-by-detection methods have become increasingly popular for real-time applications~\cite{zhang2012real} in visual tracking. The correlation filter-based trackers attract more attention in recent years due to its high speed performance \cite{chen2015experimental}. However, those conventional tracking methods~\cite{zhang2012real,zhang2014fast,danelljan2014adaptive,li2014scale,danelljan2014accurate,zhu2015collaborative,henriques2015high,sui2016Realtime,sui2015visual,li2015dual,zhang2013robust} require manual initialization with the ground truth at the first frame. Moreover, they are sensitive to the initialization variation caused by scales and position errors, and would return useless information once failed during tracking~\cite{wu2013online}.

Combining a detector with a tracker is a feasible solution for automatic initialization~\cite{andriluka2008people}. The detector, however, needs to be trained with large amount of training samples, while the prior information about the object of interest is usually not available in advance. In~\cite{mahadevan2009saliency}, Mahadevan \textsl{et al.} proposed a saliency-based discriminative tracker with automatic initialization, which builds the motion saliency map using optical flow. This technique, however, is computational intensive and not suitable for real-time applications. Some recent techniques on salient object detection and visual tracking~\cite{ma2015hierarchical,hong2015online} have achieved superior performance by using deep learning. However, these methods need large amount of samples for training. 

In~\cite{zhang2015minimum}, Zhang \textsl{et al.} proposed a fast salient object detection method based on minimum barrier distance transform. Since the saliency map effectively discovers the spatial information of the distinct objects in a scene, it enables us to improve the localization accuracy of the salient objects during tracking. In this paper, we propose a scale adaptive object tracking framework by integrating two complementary processes: salient object detection and visual tracking. A Kalman filter is employed to predict a coarse location of the object of interest. Then, a detector is used to refine the location of the object. The optimal state of the object in each frame is estimated relying on the recursive process of prediction, refining, and correction. 

\begin{figure}[tb]
	\centering
	\includegraphics[width=0.48\textwidth]{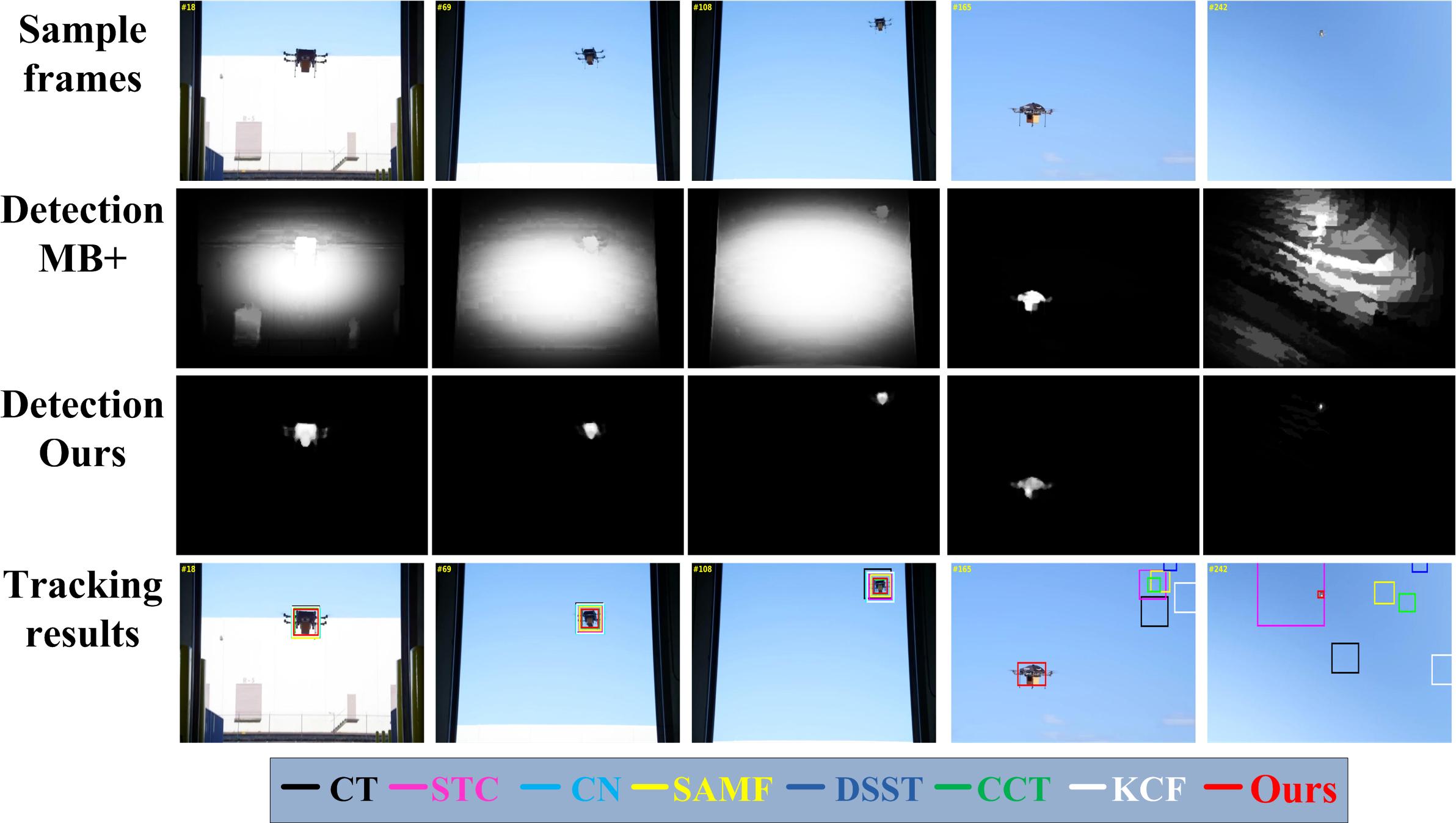} 
	\caption{A comparison of the proposed approach with the state-of-the-art detector MB+~\cite{zhang2015minimum}, and trackers CT~\cite{zhang2012real}, STC~\cite{zhang2014fast}, CN~\cite{danelljan2014adaptive}, SAMF~\cite{li2014scale}, DSST~\cite{danelljan2014accurate}, CCT~\cite{zhu2015collaborative}, KCF~\cite{henriques2015high} on a sequence~\cite{AmazonDrone2013} with challenging situations, including scale variation, out-of-view and re-appearance. For detection, MB+ fails to provide a high quality saliency map. For tracking, the existing state-of-the-art trackers cannot handle the out-of-view and re-appearance challenges. Our method provides high quality of saliency map in detection, and accurate scale and position of the target in tracking.} 
	\label{fig:Introduction}
\end{figure}

The proposed approach has been compared with the state-of-the-art detector and trackers on a sequence with challenging situations, including scale variation, rotation, illumination change, and out-of-view and re-apperance. As shown in Fig.~\ref{fig:Introduction}, for object detection, the single view saliency detection algorithm (MB+)~\cite{zhang2015minimum} is not able to provide high quality saliency maps for the image sequence because it does not manage to use the contextual information between consecutive frames; and from the tracking perspective, the existing trackers are not capable to handle out-of-view and re-appearance challenges. 

In summary, our contributions are threefold: 1) The proposed algorithm integrates a saliency map into a dynamic model and adopts the target-specific saliency map as the observation for tracking; 2) we develop a tracker with automatic initialization for a UAV sensing system; and 3) the proposed technique achieves better performance than the state-of-the-art trackers from extensive experimental evaluations.

Our remaining part of this paper is organized as follows: Some related work is briefly reviewed in section~\ref{sec:RelatedWork}; in section~\ref{sec:ProposedMethod}, the proposed approach is discussed thoroughly; in section~\ref{sec:Experiments}, we demonstrate the quantitative and qualitative evaluation results, and some limitations; and finally, the paper is concluded in section~\ref{sec:conclusions}. 

\section{Related Works}
\label{sec:RelatedWork}
Salient object detection and visual tracking plays important roles in many computer vision based applications, such as traffic monitoring, surveillance, video understanding, face recognition, and human-computer interaction~\cite{yilmaz2006object}. 

The task of salient object detection is to compute a saliency map and segment an accurate boundary of that object. For natural images, the assumptions on background and object have been shown to be effective for salient object detection~\cite{cheng2015global,zhu2014saliency}. One of the most widely used assumptions is called contrast prior, which assumes high appearance difference between the object and background. Region-based salient object detection has become increasingly popular with the development of superpixel algorithm~\cite{borji2014salient}. In addition to contrast prior and region-based methods, several recent approaches exploit boundary connectivity~\cite{zhang2015minimum,zhu2014saliency}. Wei \textsl{et al.}~\cite{wei2012geodesic} proposed a geodesic saliency detection method based on contrast, image boundary, and background priors. The salient object is extracted by finding the shortest path to the virtual background node. Zhu \textsl{et al.}~\cite{zhu2014saliency} formulated the saliency detection as an optimization problem and solved it by a combination of superpixel and background measurement. Cheng \textsl{et al.}~\cite{cheng2015global} computed the global contrast using the histogram and color statistics of input images. These state-of-the-art saliency detection methods achieve pixel-level resolution. Readers may refer to~\cite{borji2014salient} for a comprehensive review on salient object detection. 

The goal of visual tracking is to estimate the boundary and trajectory of the object in every frame of an image sequence. Designing an efficient and robust tracker is a critical issue in visual tracking, especially in challenging situations, such as illumination variation, in-plane rotation, out-of-plane rotation, scale variation, occlusion, background clutter and so on~\cite{wu2013online}. Over the past decades, various tracking algorithms have been proposed to cope with the challenges in visual tracking. According to the models adopted, these approaches can be generally classified into generative models~\cite{ross2008incremental,sui2015robust,yin2011hierarchical,yugroupwise2016}, and discriminative models~\cite{zhang2012real,mahadevan2009saliency,kalal2012tracking}. Ross \textsl{et al.}~\cite{ross2008incremental} exploited an incremental subspace learning to visual tracking, which assumes that the obtained temporal targets reside in a low-dimensional subspace. Sui \textsl{et al.}~\cite{sui2015robust} proposed a sparsity-induced subspace learning which selects effective features to construct the target subspace. Yin \textsl{et al.}~\cite{yin2011hierarchical} proposed a hierarchical tracking method based on the subspace representation and Kalman filter. Yu \textsl{et al.}~\cite{yugroupwise2016} introduced a large-scale fiber tracking approach based on Kalman filter and group-wise thin-plate spline point matching. 

The discriminative tracking-by-detection approaches have become increasingly popular in recent years. Zhang \textsl{et al.}~\cite{zhang2012real} proposed a real-time tracker based on compressive sensing. Mahadevan \textsl{et al.}~\cite{mahadevan2009saliency} proposed a saliency-based discriminative tracker, which learns the salient features based on Bayesian framework. Kalal \textsl{et al.}~\cite{kalal2012tracking} introduced a long-term tracker which enables a re-initialization in case of tracking failures.

In particular, the correlation filter-based discriminative tracking methods have attracted much attention and achieved significant progress~\cite{chen2015experimental}. Henriques \textsl{et al.}~\cite{henriques2015high} proposed a tracker using kernelized correlation filters (KCF). Zhu \textsl{et al.}~\cite{zhu2015collaborative} extended the KCF to a multi-scale kernelized tracker in order to deal with the scale variation. Zhang \textsl{et al.}~\cite{zhang2014fast} proposed a tracker via dense spatio-temporal context learning. Danelljan \textsl{et al.}~\cite{danelljan2014accurate} introduced a discriminative tracker using a scale pyramid representation. Li \textsl{et al.}~\cite{li2014scale} proposed to tackle the scale variation by integrating different low-level features. Danelljan \textsl{et al.}~\cite{danelljan2014adaptive} designed a tracker by adaptive extension of color attributes. Readers can refer to~\cite{yilmaz2006object,smeulders2014visual} and the references therein for details about visual tracking.  

\section{The Proposed Approach}
\label{sec:ProposedMethod}
\begin{figure}[tb]
	\centering
	\includegraphics[width=0.4\textwidth]{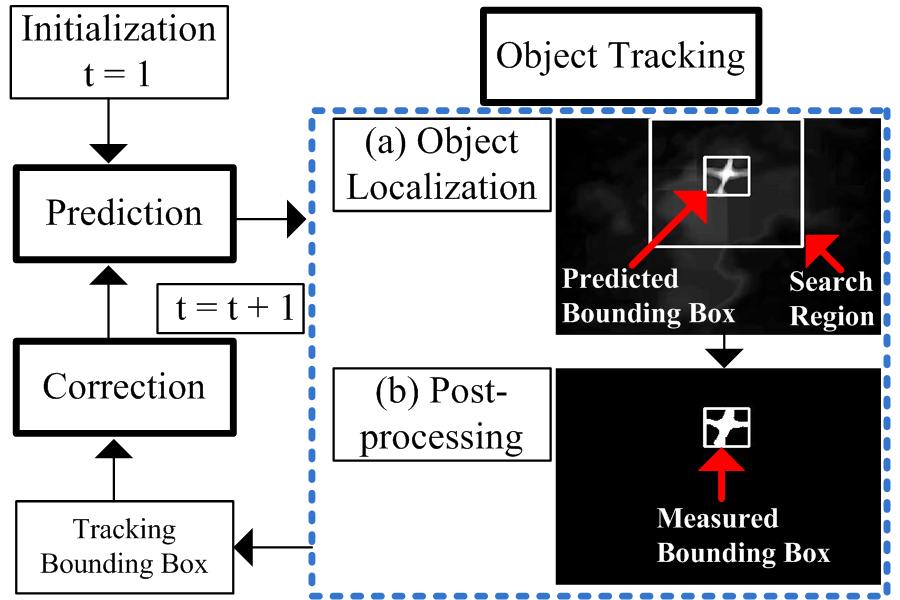} 
	\caption{The flow-chart of the proposed fast object localization and tracking strategy.}
	\label{fig:framework}
\end{figure}

The proposed fast object localization and tracking (FOLT) algorithm is formulated within a robust Kalman filter framework~\cite{yin2011hierarchical,weng2006video} to estimate the optimal state of the salient object from the saliency map in every frame of a given image sequence. Our tracking approach relies on a recursive process of prediction, object detection, and correction, as shown in Fig.~\ref{fig:framework}. A linear dynamic (with constant velocity) model has been employed to represent the transition of the motion state of the salient object in a scene~\cite{weng2006video}. The tracker is initialized on the first frame using the saliency map computed on the entire image. The motion state is predicted on each frame according to the motion states of previously obtained object of interest. Under the constraint of natural scenes, the prediction is not far away from the ground truth~\cite{yin2011hierarchical}, however, it only provides a coarse state estimation (``predicted bounding box'' shown in Fig.~\ref{fig:framework}) about the target location. We take this predicted coarse location as an initial region for further estimation during object tracking. The refined target is marked by a bounding box on that frame according to its motion state. Finally, the Kalman gain and a posteriori error covariance in the dynamic model are updated. The details about prediction, object localization and correction scheme are discussed in the following.

\subsection{Dynamic model formulation}
\label{sec:Dynamicmodel}
In the dynamic model, the object of interest is defined by a motion state variable $s = \{x,y,u,v,w,h\}$, where $(x, y)$ denotes the center coordinates, $(u, v)$ denotes its velocities, and $(w, h)$ denotes its width and height. The state at each frame $t+1$ is estimated using a linear stochastic difference equation $s(t+1) = H s(t) + u(t) + w(t)$, where the prediction noise $w(t) \sim N(0, Q)$ is Gaussian distributed with covariance $Q$, $u(t) \in \mathcal{R}^{l}$ is a driving function with dimension $l$. The vector $s \in \mathcal{R}^{n}$ describes the motion  states of the salient object. The orthogonal transition matrix $H \in n\times n$ evolves the state from the previous frame $t$ to the state at the current frame $t+1$. The vector $y(t) \in \mathcal{R}^m$ is denoted as an observation or measurement with dimension $m$ measured in frame $t$. In our notation, we will define $s_t \equiv s(t)$, and $y_t \equiv y(t)$. With the driving function removed from our model, the autoregressive model of the salient object in a frame is built based on the following linear stochastic model 
\begin{equation}
	s_{t+1} = H s_t + w_t, w_t \sim N(0, Q)\,,
	\label{equ:TransitionModel}
\end{equation}
\begin{equation}
	y_{t} = C s_t + v_t, v_t \sim N(0,R)\,,
	\label{equ:MeasureModel}
\end{equation}
where the measurement matrix $C$ is $m \times m$, and the measurement noise $v_t \sim N(0,R)$ is Gaussian distributed with covariance $R$. The diagonal elements in the prediction noise covariance matrix $Q$ and measuremnt noise matrix $R$ represent the covariance of the size and position. 

In the $t$-th frame, given the probability of $y_t$ and all the previously obtained states from the first to $(t-1)$-th frame, denoted as $s_{1:t-1}$, the optimal motion state of the target in the $t$-th frame, denoted as $s_t$, is predicted by maximizing the posterior probability $p(s_t|y_t,s_{1:t-1})$. To simplify our model, we inherit the Markov assumption, which states that the current state is only dependent on the previous state. Therefore, the objective function becomes  
\begin{equation}
	\hat{s}_t = \arg \underset{s}{max} \quad p(s_t|y_t,s_{t-1})\,.
	\label{equ:posteriori1}
\end{equation}
Using Bayes formula, the posterior probability becomes:
\begin{equation}
	p(s_t|y_t, s_{t-1}) = \frac{p(y_t|s_t, s_{t-1})p(s_t|s_{t-1})}{p(y_t)}\,,
	\label{equ:posteriori2}
\end{equation}
where the denominator $p(y_t)$ is a normalization constant, which indicates the prior distribution of the observation $y_t$. The observation $y_t$ is not dependent on the previous state $s_{t-1}$ in the presence of $s_t$, since $y_t$ is only generated by the state $s_t$. Thus equation~(\ref{equ:posteriori2}) is reduced to:
\begin{equation}
	p(s_t|y_t,s_{t-1})=p(y_t|s_t)p(s_t|s_{t-1})\,,
	\label{equ:posteriori3}
\end{equation}
where the observation model $p(y_t|s_t)$ measures the likelihood to be the target of the measurement $y_t$ with the motion state $s_t$. Finally, we formulate the objective function as 
\begin{equation}
	\hat{s}_t = \arg \underset{s}{max} \quad p(y_t|s_t)p(s_t|s_{t-1})\,.
	\label{equ:posterior4}
\end{equation}
The state estimation can be converted to the standard Kalman filter framework~\cite{KF1995} when we assumed that the state transition model and the observation model follow a Gaussian distribution.

\subsection{Object tracking}
\label{sec:Objecttracking}
\begin{figure}[tb]
	\centering
	\includegraphics[width=0.44\textwidth]{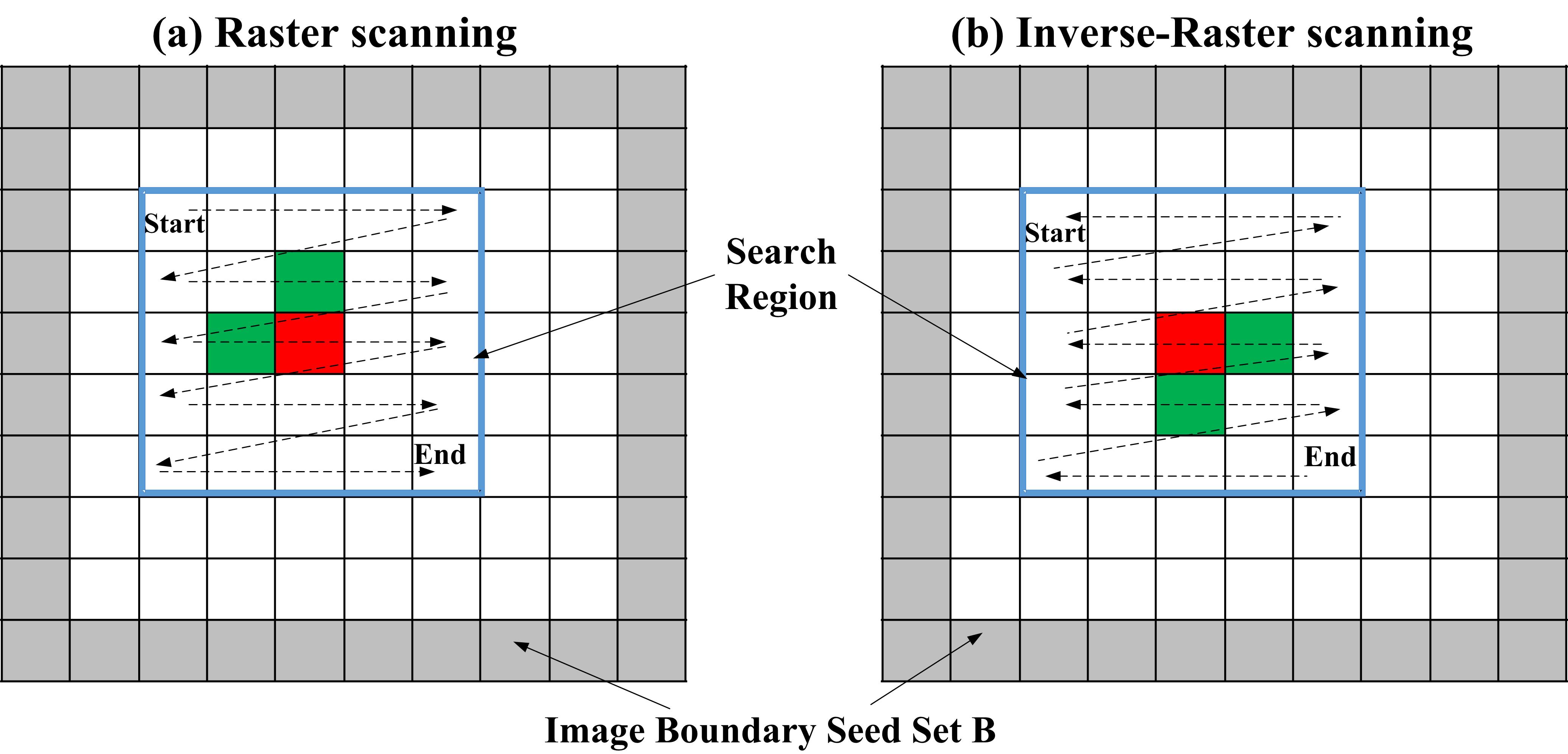} 
	\caption{Illustration of updating the search region on a 4-adjacent graph using 
		(a) Raster scanning, to update the intensity value at pixel $z$ (red box) using its two adjacent neighbors (green boxes) in search region from ``Start'' to ``End'' line by line, (b) Inverse-raster scanning, using the same updating strategy in opposite direction from ''End'' to ''Start''.}
	\label{fig:Rasterscan}
\end{figure}
The step of object tracking consists of two procedures: to implement object localization in the search region; and to infer the target state after conducting the post-processing on the saliency map. 
\subsubsection{Object localization}
\label{sec:salientlocalization}
A background prior, i.e., the image boundary connectivity cue~\cite{zhang2015minimum}, has been applied to locate the object in each frame. However, the proposed localization method has two obvious differences. First, by integrating with the contextual information, the proposed approach is capable to localize the salient object in both individual images and video sequences. Second, to leverage this cue, the saliency map, which represents the probability of a certain region in an image to be a salient object or background, is updated locally based on the coarse prediction (``predicted bounding box'' shown in Fig.~\ref{fig:Rasterscan}). 

In this paper, we will denote $I_t$ as the image on the $t$-th frame. Firstly, on the frame $t$, under the constrain of natural image~\cite{weng2006video}, it is reasonable to define a search region, where the salient object is guaranteed to exist, by expanding the predicted bounding box with certain percentage $\delta$. Next, the saliency map $D$ in the search region is updated by computing the minimum barrier distance~\cite{zhang2015minimum,strand2013minimum} with respect to a set of background seed set pixels $B$ (see illustration in Fig.~\ref{fig:Rasterscan}). While the values for the pixels that are not in the search region are kept the same. Through those two steps, the position and scaling of the object is estimated on the frame $t$.

It is assumed under the image background connectivity cue that background regions are normally connected to the baground seed set $B$. In this paper, a path $p$ from pixel $1$ to pixel $n$ consists of a sequence of pixels, and is denoted as $p = \langle p(1), p(2), \cdots, p(n) \rangle$. In this sequence, each of the two consecutive pixels are neighbors. Each pixel in a 2-D single-channel image $I_t$ is denoted as a vertex. The neighboring pixels are connected by edges. In this work, we consider 4-adjacent neighbors as demonstrated in Fig.~\ref{fig:Rasterscan}. For the image $I_t$, the cost function of computing the distance of a path from pixel $z$ to the background seed set $B$ is defined as finding the difference between the maximum and minimum intensity values in this path. The formula of the cost function is  
\begin{equation}
	F(p) =\max_{j=1}^{n} I_t(p(j)) - \min_{j=1}^{n} I_t(p(j))\,, 
	\label{equ:MBDcostfunc1}
\end{equation}
where $I_t(\cdot)$ denotes the intensity value of a pixel in frame $t$. The saliency map $D(z)$ is obtained by minimizing the cost function $F(p)$,
\begin{equation}
	D(z) = \underset{p \in \Theta_{B,z}}{min} F(p)\,,
	\label{equ:MBDcostfunc2}
\end{equation}
where $p \in \Theta_{B,z}$ denotes the set, which includes all the possible paths from pixel $z$ to the background seed set $B$. This formulates the computation of the saliency map as a problem of finding the shortest path for each pixel in the image $I_t$. It can be solved by scanning each pixel using the Dijkstra-like algorithm. We denote $\langle m,z \rangle$ as the edge between two connected pixels $m$ and $z$, and $p(m)$ as the path asigned to pixel $m$, and $p_m(z)$ as the path connected pixel $m$ and pixel $z$ with edge $\langle m,z \rangle$. Therefore, the cost function of $p_m(z)$ is evaluated using  
\begin{equation}
	\begin{split}
		F(p_m(z)) = & \max(U_t(m), I_t(z)) -\\ & \min(L_t(m), I_t(z))\,,
	\end{split}
	\label{equ:MBDcostfunc3}
\end{equation}
where $U_t(m)$ and $L_t(m)$ are matrices with the highest and lowest pixel values for the path $p(m)$, respectively. The initial values of $U$ and $L$ are identical with the image $I$. In the initial saliency map $D$, the region corresponding to the image boundary seed set $B$ is initialized with intensity of zeros, and the left pixels are initialized with intensity of infinity. The raster scanning pass and inverse-raster scanning pass are implemented alternately to update the saliency map $D$, as shown in Fig.~\ref{fig:Rasterscan}. The number of passes is denoted as $N$, and in this paper we select $N=3$. In the raster scanning pass, each pixel in the search region is visited line by line. The intensity value at pixel $z$ is updated using its two neighbors (as illustrated in Fig.~\ref{fig:Rasterscan}). The inverse-raster scanning applies the same updating but in an opposite order. The updating strategy of saliency map $D(z)$ is 
\begin{equation}
	D(z) \leftarrow \min \left\{
	\begin{array}{l}
		D(z) \\
		F(p_m(z))
	\end{array}
	\right.
	\label{equ:MBDupdate}
	\,.
\end{equation}

\subsubsection{Post-processing}
\label{sec:postprocessing}
\begin{figure}[tb]
	\centering
	\includegraphics[width=0.48\textwidth]{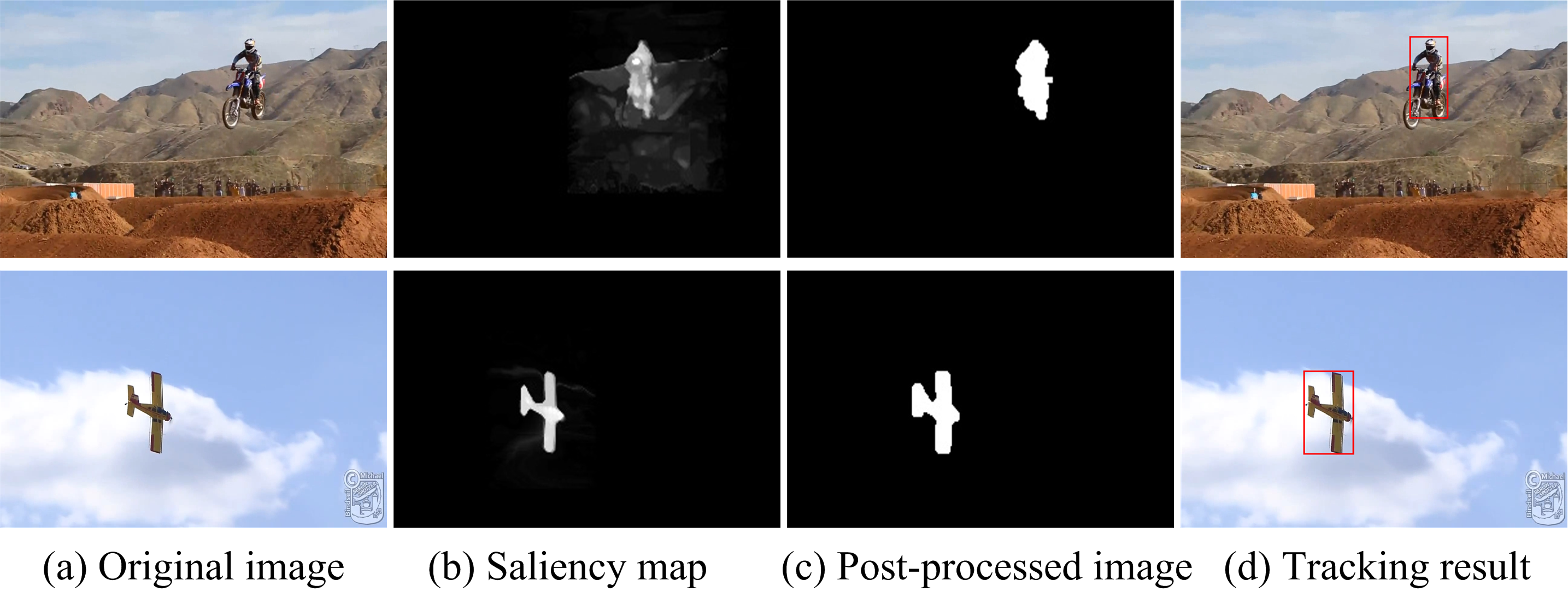}
	\caption{Illustration of the intermediate processes, including salient object detection, localization and post-processing in the proposed tracking approach. The two image sequences are \textsl{motorcycle\_011} (frist row)~\cite{li2015nus} and \textsl{airplane\_016} (second row)~\cite{li2015nus}.}
	\label{fig:Postprocessing}
\end{figure}
Two efficient post-processing operations have been implemented to improve the quality of the saliency map, and to further segment the salient object in every frame. A Threshold is applied to the saliency map, which transforms the saliency map to a binary image. Then, the tracking bounding box is extracted after dilation (see Fig.~\ref{fig:framework}). Global threshold is not an efficient solution in the scenario where image has non-uniform illumination and lighting conditions. Hence, it is wise to employ adaptive threshold~\cite{GonzalezDIP2006}. We denote $m_{ab}$ as the mean value of the set of pixels contained in a neighboring block, $\Gamma_{ab}$, centered at coordinates $(a,b)$ in an image. The size of the block $\Gamma_{ab}$ is $\alpha \times \alpha$. The following formula defines the local thresholds 
\begin{equation}
	\label{equ:Threshold1}
	T_{ab} = m_{ab} - \lambda\,,
\end{equation}
where $\lambda$ is a nonnegative offset. The segmented image is computed as 
\begin{equation}
	f(a,b)= \left\{ 
	\begin{array}{ll}
		$1$, \quad \text{if $g(a,b)$} \geq T_{ab}\\
		$0$, \quad \text{otherwise}
	\end{array}
	\right.
	\label{equ:Threshold2}
	\,,
\end{equation}
where $g(a,b)$ is the input saliency map image. The equ.~(\ref{equ:Threshold2}) is evaluated for all pixels in the image, and a threshold is calculated at each position $(a,b)$ using the pixels in the neighboring block of $\Gamma_{ab}$. The idea of dilation is applied to enhance the quality of the thresholded image $f(a,b)$. The dilation of $A$ by $S$, denoted by $A \oplus S$, is defined as 
\begin{equation}
	A \oplus S = \{q|\hat{(S)}_q \cap A \neq \varnothing \}\,,
	\label{equ:Dilation}
\end{equation}
where $S \in \mathcal{R}^{O \times M}$ is a structuring element. After dilation, the minimum bounding box of the extracted region gives the state estimation $\hat{s}_t$ of the target. Fig.~\ref{fig:Postprocessing} shows examples on two sequences (\textsl{motorcycle\_011} and \textsl{airplane\_016}). The tracking bounding box is estimated by passing the original image through the processes of salient object detection and post-processing.  
\subsection{Fast object localization tracking}
\label{sec:FOLTupdatescheme}
\begin{algorithm}[tb]
	\caption{Fast Object Localization Tracking}
	\label{alg:FOLT}
	\KwIn{image $I_t$, saliency map $D$, number of pass $N$}
	\KwOut{target state $\hat{s}_t$}
	Set $D$ to $\infty$ in search region and $B$\\ 
	Keep the values in $D$ outside the search region\\
	\For{each frame}
	{
		Prediction in Kalman filter \\
		Object tracking: \\
		\For{$i = 1 : N$}
		{
			\If{mod$(i,2) = 1$}
			{Raster Scanning using~(\ref{equ:MBDupdate})}
			\Else
			{
				Inverse-Raster Scanning using~(\ref{equ:MBDupdate})
			}
		}
		Observation: update measurement using~(\ref{equ:MeasureModel})\\
		Correction: update~(\ref{equ:KalmanGain}),~(\ref{equ:PosterrCov})\\
	}
	Compute the entire $D$ every $10$ frames
\end{algorithm}
In this section, we denote $G_t^{-}$ and $G_t$ as a priori error covariance and a posteriori error covariance, respectively. In the correction stage of frame $t$, both of the posterior error covariance and Kalman gain are updated as follows. 
\begin{enumerate}
	\item Compute Kalman gain
	\begin{equation}
		K_t = \frac{H^T}{G_t (H G_t^{-} H^T + R)}
		\label{equ:KalmanGain}
	\end{equation}
	\item Update estimate with measurement $y_t$
	\begin{equation}
		\hat{s}_t = \hat{s}_t^{-} + K_t (y_t - C \hat{s}_t^{-})
		\label{equ:EstimateKFupdate}
	\end{equation}
	\item Compute a posteriori error covariance 
	\begin{equation}
		G_t = (\mathcal{I} - K_t H) G_t^{-}
		\label{equ:PosterrCov}
	\end{equation}
\end{enumerate}

In summary, through the recursive prediction, object localization, and correction, the salient object in a image sequence is automatically detected and tracked. The details of the fast object localization tracking is illustrated in Alg.~\ref{alg:FOLT}. The saliency map $D$ is updated on the entire image every $10$ frames as a trade-off between the accuracy and the speed. 

\section{Experimental Evaluations}
\label{sec:Experiments}
\setlength{\tabcolsep}{4.5pt}
\begin{table}[tb]
	\renewcommand{\arraystretch}{1.2}
	\caption{Attributes used to characterize each image sequence from a tracking perspective. We denote scale variation as SV, illumination variation as IV, background interference as BGI, and in-plane rotation and out-of-plane rotation as IPR \& OPR.}
	\label{tab:AllImageSnapshot}
	\centering
	\begin{tabular}{|l|l|l|}
		\hline
		Image Sequence & Object of Interest & Tracking challenges \\  
		\hline
		Aircraft~\cite{mian2008realtime} & Propeller plane & SV, IPR \& OPR\\
		airplane\_001~\cite{li2015nus} & Jet plane & SV, IPR \& OPR\\
		airplane\_006~\cite{li2015nus} & Jet plane & SV, IPR \& OPR\\
		\hline
		airplane\_011~\cite{li2015nus} & Propeller plane & SV, IPR \& OPR\\
		airplane\_016~\cite{li2015nus} & Propeller plane & SV, IPR \& OPR, BGI\\
		big\_2~\cite{mian2008realtime} & Jet plane & SV, IPR \& OPR, BGI\\
		Plane\_ce2~\cite{liang2015encoding} & Jet plane & SV, IPR \& OPR, BGI\\
		\hline
		Skyjumping\_ce~\cite{liang2015encoding} & Person & SV, IV, IPR \& OPR, BGI\\
		\hline
		motorcycle\_006~\cite{li2015nus} & Person and motorcycle & SV, IPR \& OPR, BGI\\
		\hline	
		surfing~\cite{VOT2014} & Person & SV, IPR \& OPR, BGI\\
		Surfer~\cite{wu2013online} & Person & SV, IPR \& OPR, BGI\\
		Skater~\cite{wu2013online} & Person & SV, IPR \& OPR, BGI\\
		\hline
		Sylvester~\cite{wu2013online} & Doll & IV, IPR \& OPR, BGI\\	
		ball~\cite{VOT2014} & ball & SV, BGI\\
		\hline	
		Dog~\cite{wu2013online} & Dog & SV, OPR\\
		\hline			
	\end{tabular}
\end{table}
The proposed approach is implemented in C++ with OpenCV 3.0.0 on a PC with an Intel Xeon W3250 2.67 GHz CPU and 8 GB RAM. The dataset and source code of the proposed approach will be available on the author's homepage. The proposed tracker is evaluated on 15 popular image sequences collected from~\cite{li2015nus,liang2015encoding,VOT2014,wu2013online}. There are a total of over 6700 frames in the dataset. The highest and lowest frame sizes are $1280 \times 720$ and $320 \times 240$, respectively. The dataset, from tracking perspective of view, includes different scenarios with challenging situations, such as scale variation, occlusions, in-plane and out-of-plane rotations, illumination, and background interference, as shown in Table.~\ref{tab:AllImageSnapshot}. The image sequences without sky-region scenarios, i.e. \textsl{Skyjumping\_ce, motorcycle\_006, surfing, Surfer, Skater, Sylvester, ball} and \textsl{Dog} have been selected to test the robustness and generality of the proposed approach within different scenarios. In each frame of these video sequences, we labeled the target manually in a bounding box, which is used as the ground truth in the quantitative evaluations. 

In our implementation, input images are first resized so that the maximum dimension is 300 pixels. The transition state matrix $H \in \mathcal{R}^{6 \times 6}$ and measurement matrix $C \in \mathcal{R}^{4 \times 6}$ are fixed during the experiment. The diagonal values corresponding to the position (i.e., $(x,y)$) and scale (i.e., $(w,h)$) covariance in prediction noise covariance matrix $Q$ and measurement covariance matrix $R$ are set to $0.01$ and $0.1$, respectively. Some other parameters for all image sequences are as follows. The percentage value $\delta = 0.25$, the size of block $\Gamma_{ab}$ is $5 \times 5$, the offset $\lambda = 7$, and the structuring element is $S \in \mathcal{R}^{5\times 3}$. 

We compared the proposed approach with seven state-of-the-art trackers. The seven competing trackers are manually initialized at the first frame using the ground truth of the target object. Once initialized, the trackers automatically track the target object in the remaining frames. However, the proposed tracker automatically runs to track the target object from the first frame to the end frame. Three experiments are designed to evaluate trackers as discussed in~\cite{wu2013online}: one pass evaluation (OPE), temporal robustness evaluation (TRE), and spatial robustness evaluation. For TRE, we randomly select the starting frame and run a tracker to the end of the sequence. Spatial robustness evaluation initializes the bounding box in the first frame by shifting or scaling. As discussed in Section~\ref{sec:ProposedMethod}, the proposed method manages to automatically initialize the tracker and is not sensitive to spatial fluctuation. Therefore, we applied one pass evaluation and temporal robustness evaluation in this section using the same temporal randomization as in~\cite{wu2013online}, and readers may refer to~\cite{wu2013online} for more details. 
\setlength{\tabcolsep}{1.0pt}
\begin{table}[tb]
	\renewcommand{\arraystretch}{1.1}
	\caption{Quantitative evaluations of the proposed and the seven competing trackers on the 15 sequences. The best and second best results are highlighted in bold-face and underline fonts, respectively.}
	\label{tab:AllTrackers}
	\centering
	\begin{tabular}{|l|c|c|c|c|c|c|c|c|}
		\hline
		& Ours & CT & STC & CN & SAMF & DSST & CCT & KCF \\   
		&  & \cite{zhang2012real} & \cite{zhang2014fast} & \cite{danelljan2014adaptive} & \cite{li2014scale} & \cite{danelljan2014accurate} & \cite{zhu2015collaborative} & \cite{henriques2015high} \\ 
		\hline
		Precision of TRE & \textbf{0.79} & 0.51 & 0.59 & 0.64 & 0.65 & 0.65 & \underline{0.66} & 0.60\\   
		\hline
		Success rate of TRE & \textbf{0.61} & 0.45 & 0.46 & 0.54 & \underline{0.58} & 0.56 & 0.57 & 0.52 \\  
		\hline
		Precision of OPE & \textbf{0.83} & 0.44 & 0.48 & 0.44 & 0.59 & 0.48 & \underline{0.66} & 0.48 \\  
		\hline
		Success rate of OPE & \textbf{0.66} & 0.34 & 0.41 & 0.42 & 0.52 & 0.44 & \underline{0.53} & 0.38 \\   	  \hline
		CLE (in pixel) & \textbf{14.5} & 74.4 & 38.0 & 55.0 & 40.8 & 55.7 & \underline{23.2} & 45.6 \\   
		\hline
		Average speed (in fps) & \underline{141.3} & 12.0 & 73.6 & 87.1 & 12.9 & 20.8 & 21.3 & \textbf{144.8} \\  
		\hline
	\end{tabular}
\end{table}
\subsection{Speed performance}
\label{sec:Speedperformance}
For salient object detection, the most up-to-date fast detector MB+~\cite{zhang2015minimum} attains a speed of 49 frames-per-second (fps). In contrast, the proposed method achieves a speed of 149 fps, three times faster than MB+, and the detection performance is better than MB+. For object tracking, the average speed comparison of the proposed and the seven state-of-the-art competing trackers is tabulated in Table~\ref{tab:AllTrackers}. The average speed of our tracker is 141 fps, which is at the same level as the fastest tracker KCF~\cite{henriques2015high}, however, KCF adopts a fixed tracking box, which could not reflect the scale changes of the target object during tracking. On average, our method is more than ten times faster than CT~\cite{zhang2012real} and SAMF~\cite{li2014scale}, five times faster than DSST~\cite{danelljan2014accurate} and CCT~\cite{zhu2015collaborative} and about two times faster than STC~\cite{zhang2014fast} and CN~\cite{danelljan2014adaptive}.
\begin{figure}[tb]
	\centering
	\includegraphics[width=0.24\textwidth]{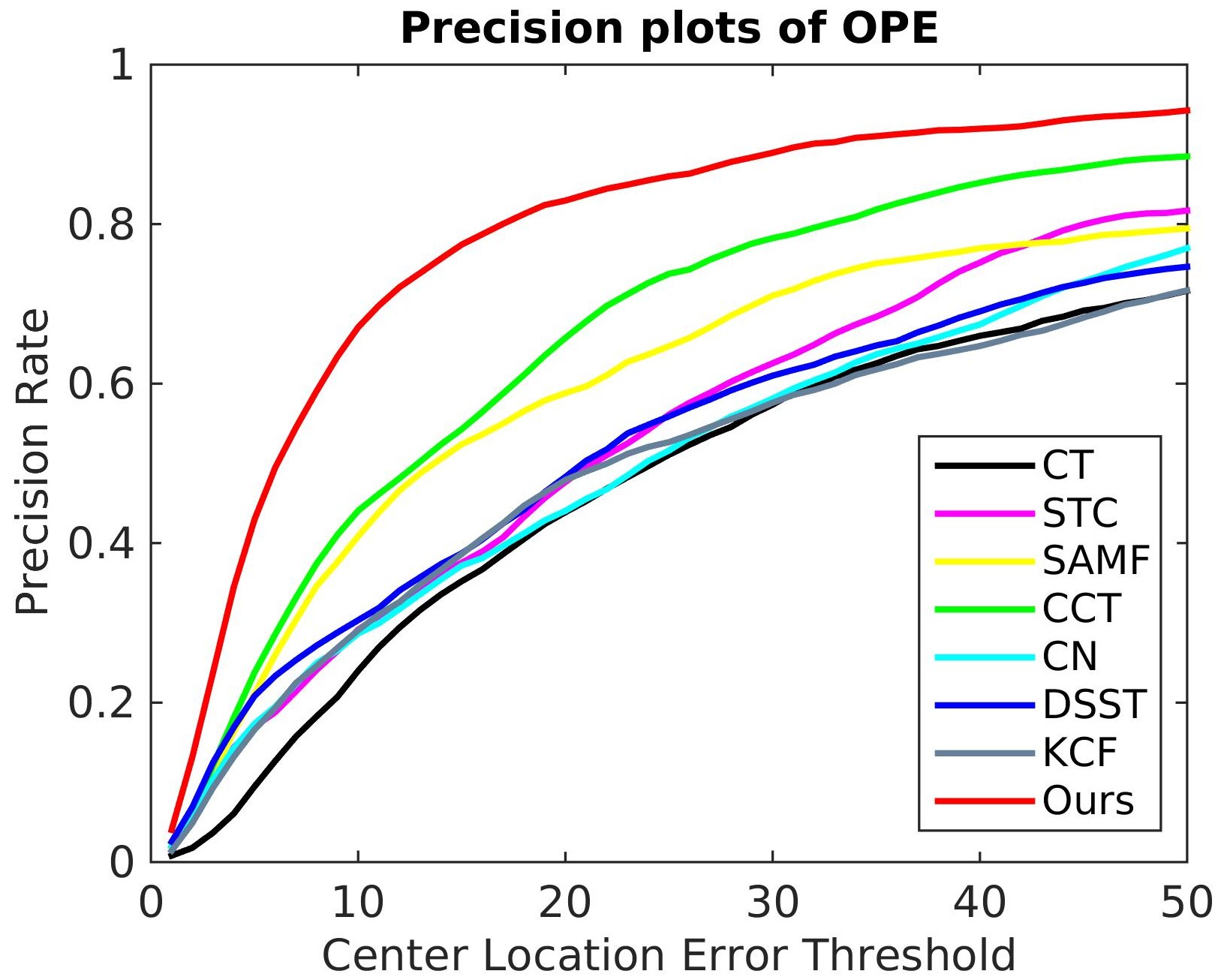} 
	\includegraphics[width=0.24\textwidth]{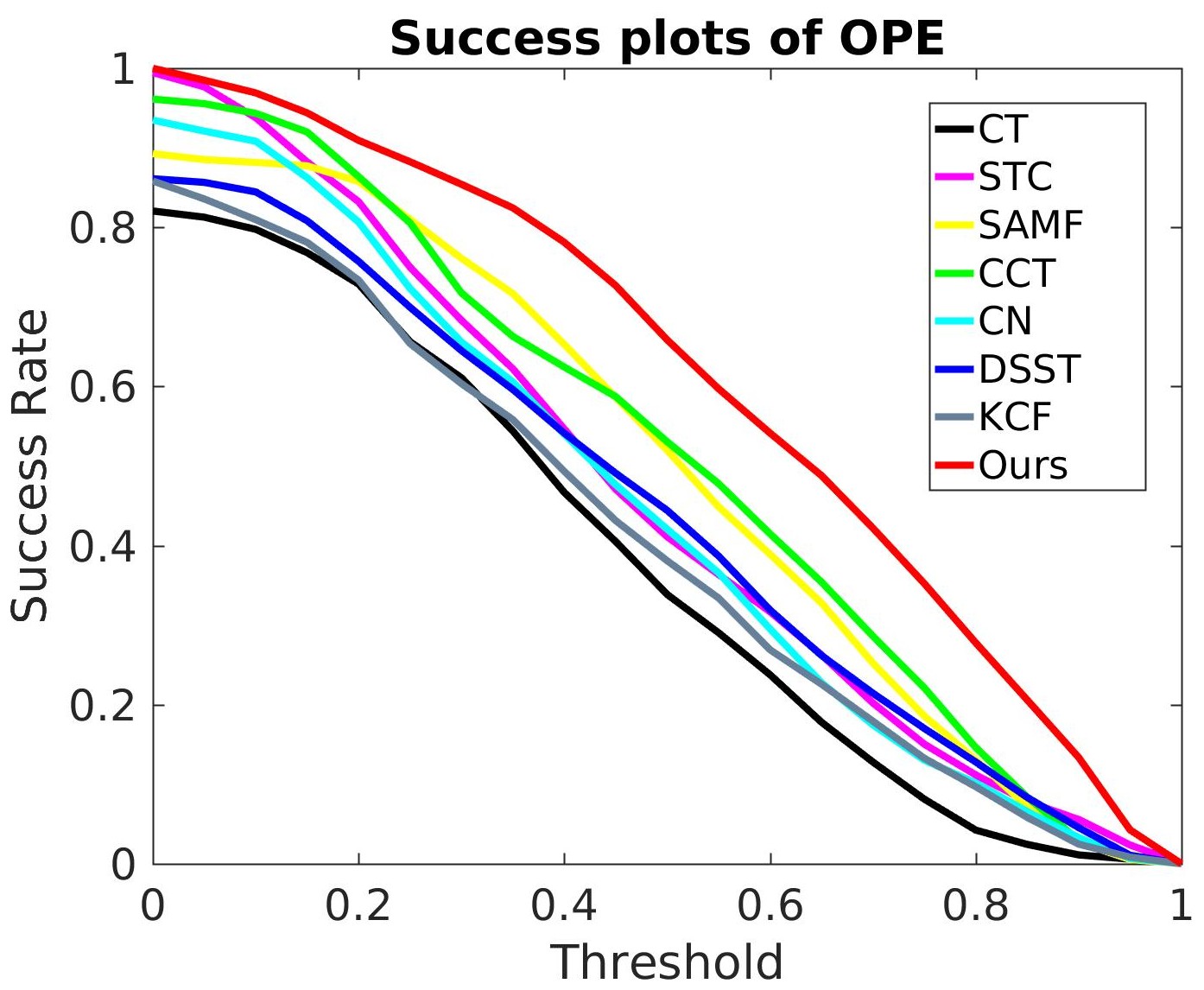} 
	\includegraphics[width=0.24\textwidth]{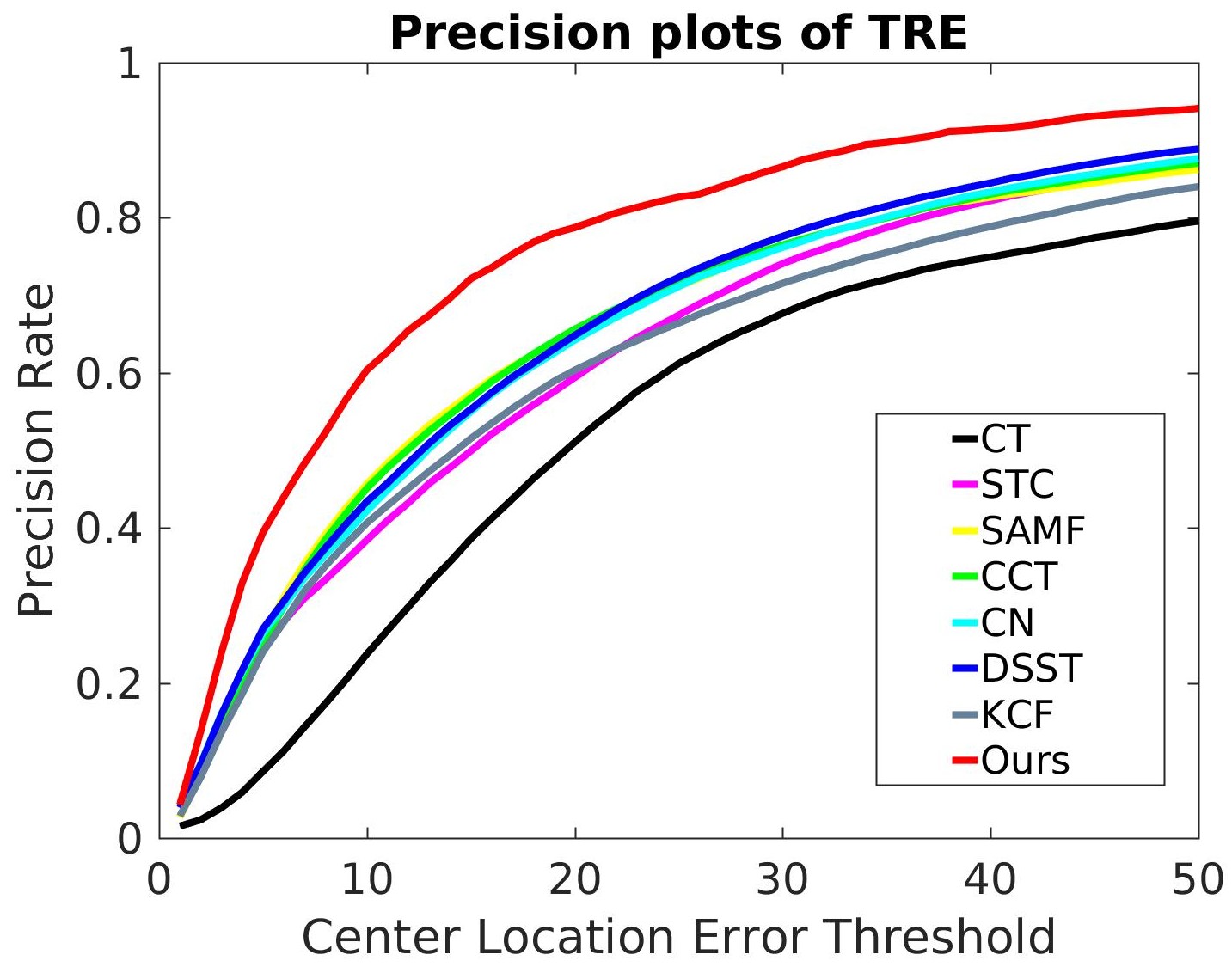} 
	\includegraphics[width=0.24\textwidth]{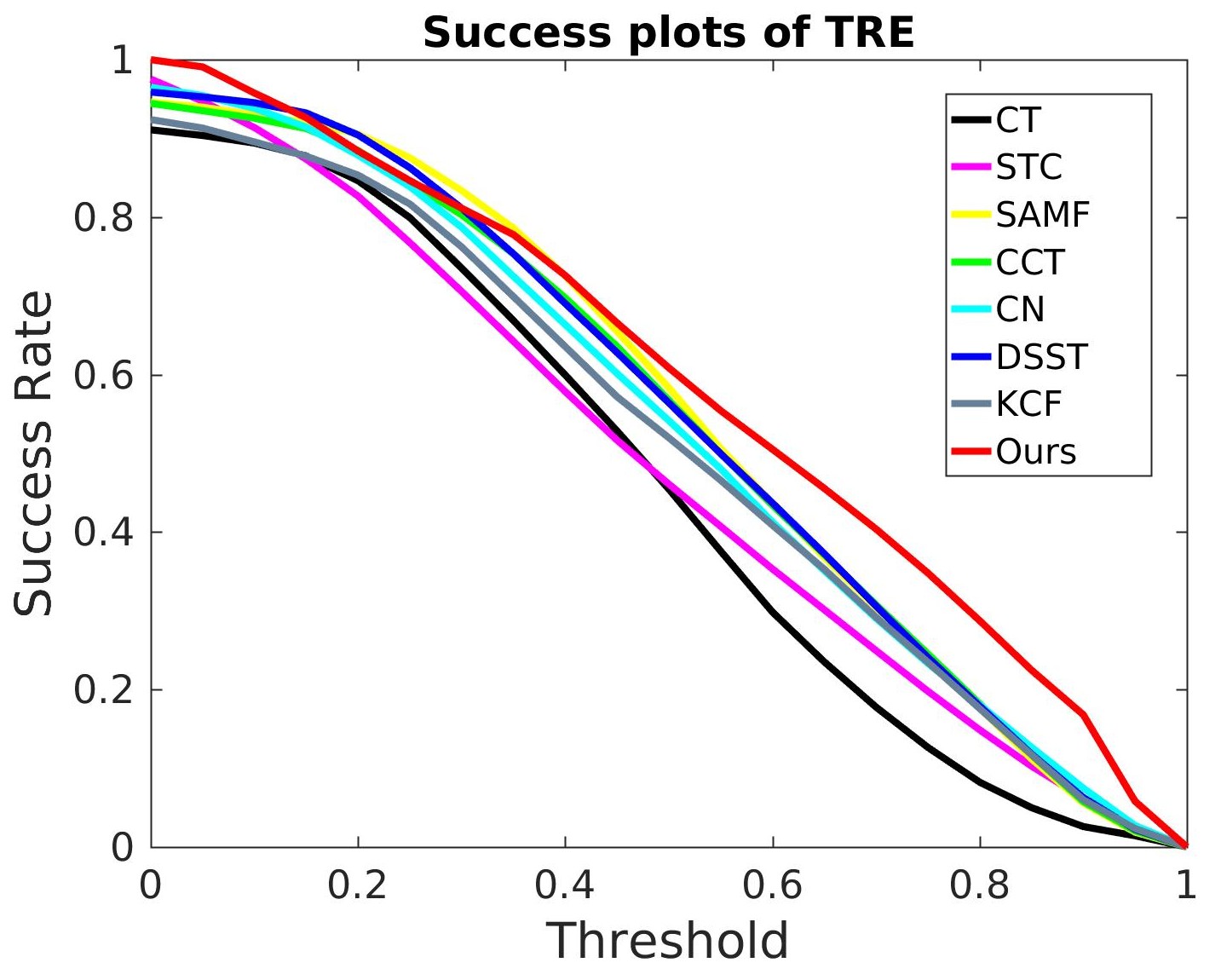} 
	\caption{Average Precision and success rate plots over the 15 sequences in (top) one pass evaluation (OPE) and (bottom) temporal robustness evaluation (TRE). (best viewed in color)}
	\label{fig:PRSRPlot}
\end{figure}
\subsection{Comparison with the state-of-the-art trackers}
\label{sec:CompareAllTrackers}
The performance of our approach is quantitatively evaluated following the metrics used in~\cite{wu2013online}. We present the results using precision, centre location error (CLE), and success rate (SR). The CLE is defined as the Euclidean distance between the centers of the tracking and the ground-truth bounding boxes. The precision is computed from the percentage of frames where the CLEs are smaller than a threshold. Following~\cite{wu2013online}, the threshold value is set at 20 pixels for the precision in our evaluations. A tracking result in a frame is considered successful if $\frac{a_t\bigcap a_g}{a_t\bigcup a_g}>\theta$ for a threshold $\theta\in\left[0,1\right]$, where $a_t$ and $a_g$ denote the areas of the bounding boxes of the tracking and the ground truth, respectively. Thus, SR is defined as the percentage of frames where the overlap rates are greater than a threshold $\theta$. Normally, the threshold $\theta$ is set to 0.5. We evaluate the proposed method by comparing to the seven state-of-the-art trackers: CT, STC, CN, SAMF, DSST, CCT, and KCF.   

The comparison results on the 15 sequences are shown in Table~\ref{tab:AllTrackers}. We present the results under one-pass evaluation and temporal robustness evaluation using the average precision, success rate, and CLE over all sequences. As shown in the table, the proposed method outperforms all seven competing trackers. It is evident that, in the one pass evaluations, the proposed tracker obtains the best performance in the CLE (14.5 pixels), and the precision (0.83), which are 8.7 pixels and $17\%$ superior to the second best tracker, the CCT tracker (23.2 pixels in CLE and 0.66 in precision). Meanwhile, in the success rate, the proposed tracker achieves the best result, which is a $13\%$ improvement against the second best tracker, the SAMF tracker. Please note that, for the seven competing trackers, the average performance in TRE is higher than that in OPE; while for the proposed tracker, the average precision and success rates in TRE are lower than those in OPE. One possible reason is that the proposed tracker tends to perform well in longer sequences, while the seven competing trackers work better in shorter sequences~\cite{wu2013online}. 
\setlength{\tabcolsep}{1.8pt}
\begin{table}[tb]
	\renewcommand{\arraystretch}{1.2}
	\caption{Precision on the 15 sequences of the proposed and the seven competing trackers. The best and the second best results are highlighted in bold-face and underline fonts, respectively.}
	\label{Table:PrecisionOPE}
	\centering
	\begin{tabular}{lcccccccc}
		\hline
		& Ours &CT &STC &CN &SAMF &DSST &CCT &KCF \\ 
		\hline
		Aircraft &\textbf{0.95} &0.28 &0.48 &0.16 &0.59 &0.16 &\underline{0.71} &0.48 \\   \hline
		airplane\_001 &\textbf{0.94} &0.37 &0.25 &0.21 &0.40 &0.21 &\underline{0.50} &0.02 \\ \hline
		airplane\_006 &\textbf{0.96} &0.23 &0.44 &0.33 &\underline{0.75} &0.43 &0.64 &0.60 \\ \hline
		airplane\_011 &\textbf{1.0} &0.65 &0.30 &0.30 &0.30 &0.30 &\underline{0.74} &0.30 \\ \hline
		airplane\_016 &\textbf{0.94} &0.01 &0.68 &\underline{0.91} &0.52 &0.81 &0.79 &0.88 \\ \hline
		big\_2 &\textbf{1.0} &0.14 &\underline{0.96} &0.95 &0.93 &0.96 &0.95 &0.63 \\   \hline
		Skyjumping\_ce &\textbf{0.94} &\underline{0.93} &0.24 &0.10 &0.72 &0.31 &0.27 &0.17 \\ \hline
		Plane\_ce2 &0.82 &0.09 &0.32 &0.09 &\textbf{0.92} &0.39 &0.80 &\underline{0.86} \\ \hline 
		motorcycle\_006 &\textbf{0.47} &0.24 &\underline{0.30} &0.19 &0.14 &0.16 &0.10 &0.14 \\ \hline          
		surfing &\underline{0.94} &\textbf{1.00} &\textbf{1.00} &\textbf{1.00} &\textbf{1.00} &\textbf{1.00} &\textbf{1.00} &\textbf{1.00} \\ \hline   
		Skater &\underline{0.54} &0.02 &0.27 &\textbf{0.58} &0.44 &0.46 &0.49 &0.43 \\ \hline
		Surfer &\textbf{0.59} &0.35 &0.48 &\underline{0.51} &0.43 &0.34 &0.46 &0.22 \\ \hline             
		Sylvester &\underline{0.87} &0.62 &0.59 &\textbf{0.93} &0.85 &0.84 &0.84 &0.84 \\ \hline
		ball &\underline{0.85} &\textbf{1.00} &0.29 &0.14 &0.31 &0.19 &\textbf{1.00} &0.27 \\ \hline       
		Dog &0.63 &\underline{0.68} &0.57 &0.22 &0.54 &\textbf{0.69} &0.56 &0.35 \\ \hline
		
		Average precision rate &\textbf{0.83} &0.44 &0.48 &0.44 &0.59 &0.48 &\underline{0.66} &0.48 \\ \hline
	\end{tabular}
\end{table}

We also report the comparison results in the one pass evaluation against the
seven competing trackers on all 15 video sequences in Table~\ref{Table:PrecisionOPE} and Table~\ref{Table:SuccessOPE}, respectively.
Our approach obtains the best or the second best performance of 14 in precision
and 9 in success rate out of the 15 sequences. Fig.~\ref{fig:PRSRPlot} plots the average precision and success plots in the one pass evaluation and temporal robustness evaluation over all 15 sequences. In the two evaluations, according to both the precision and the success rate, our approach significantly outperforms the seven competing trackers. In summary, the precision plot demonstrates that our approach is superior in robustness compared to its counterparts in the experiments; the success rate shows that our method estimates the scale changes of the target more accurately.

\subsection{Qualitative evaluation}
\label{sec:QualitativeEval}
In this section, we present some qualitative comparisons of our approach with respect to the seven competing trackers. The proposed approach is generic and can be applied to track any object of interest, including non-rigid and articulated objects. In this section, we present qualitative results our tracker using eight representative image sequences to demonstrate the effectiveness using the dataset described in previous section. We assume the target object is in low resolution when more than one ground truth bounding box has less than 400 pixels. The eight image sequences are categorized into four groups based on their scenarios and tracking challenges, as shown in Table~\ref{tab:AllImageSnapshot}. 

\begin{figure}[tb]
	\centering
	\includegraphics[width=0.48\textwidth]{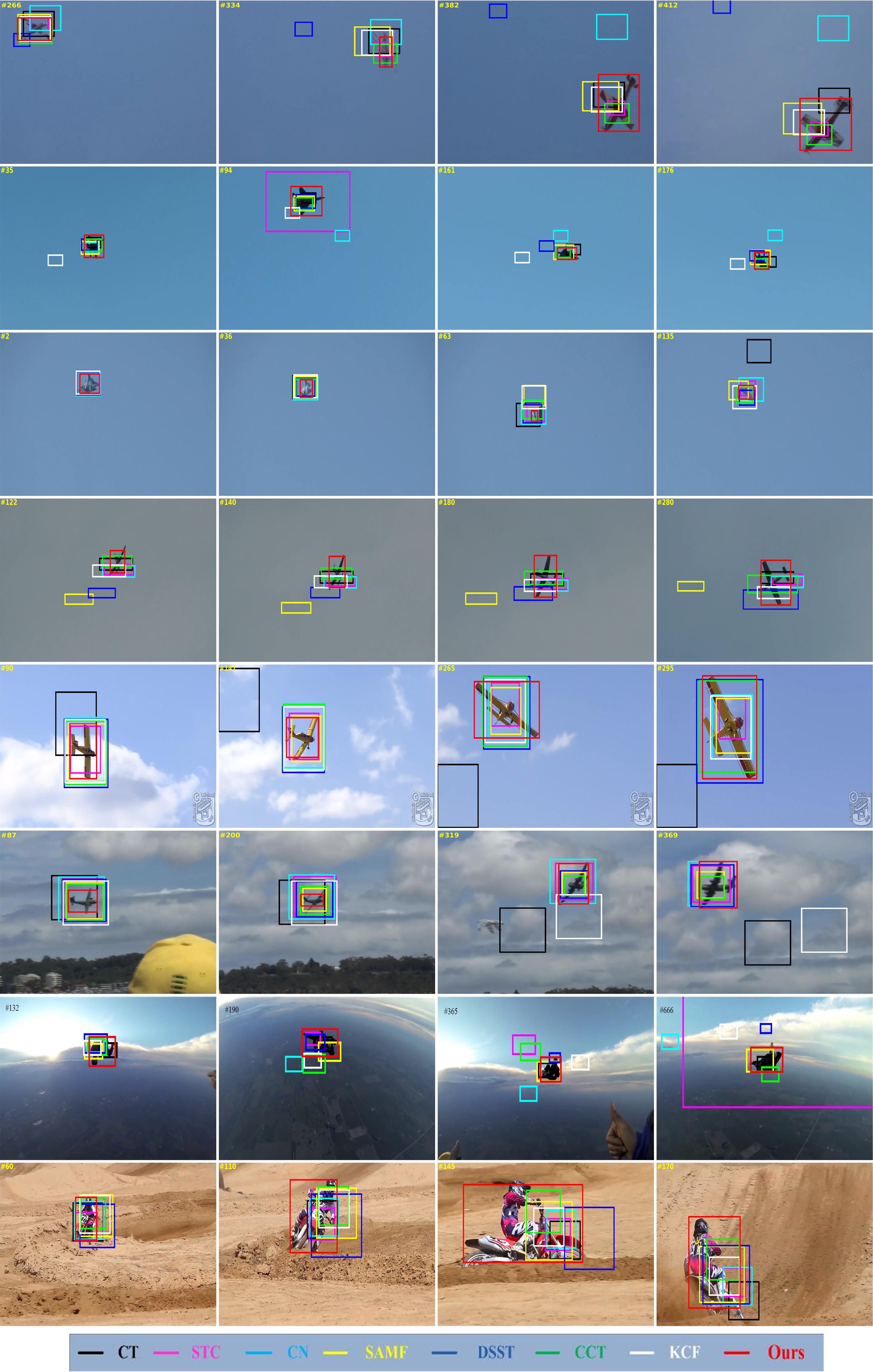} 

	\caption{Tracking results in representative frames of the proposed and the $7$ competing trackers on eight sequences with different tracking challenges and scenarios. The first row: \textsl{Aircraft~\cite{mian2008realtime}}; the second row: \textsl{airplane\_001~\cite{li2015nus}}; the third row: \textsl{airplane\_006~\cite{li2015nus}}; the fourth row: \textsl{airplane\_011~\cite{li2015nus}}; the fifth row: \textsl{airplane\_016~\cite{li2015nus}}; the sixth row: \textsl{big\_2~\cite{mian2008realtime}}; the seventh row: \textsl{Skyjumping\_ce~\cite{liang2015encoding}}; The eighth row: \textsl{motorcycle\_006~\cite{li2015nus}}. (best viewed in color)}
	\label{fig:ImgseqChallenges}
\end{figure}

The first group has clear sky as the background, including three image sequences, \textsl{Aircraft}, \textsl{airplane\_001} and \textsl{airplane\_006}, which are shown from the first row to the third row in Fig.~\ref{fig:ImgseqChallenges}. All the three image sequences have scale variation and in-plane and out-of-plane rotations. The propeller plane in image sequence \textsl{Aircraft} are in high resolution. The jet planes in the other two image sequences are in low resolution, which increases the difficulity of tracking. Moreover, the background near the jet plane in the image sequence \textsl{Airplane\_006} has an appearance similar to the target. The competing trackers STC, SAMF, DSST, and CCT were proposed to deal with scale variation, but they failed in the three image sequences. The predicted bounding box is either too large or too small. The reason is that their scaling strategy depends on the hard-coded scaling ratio, which is not adaptive to rotation and scale variation of the target. In contrast, our tracker is based on a saliency map, which leads to an accurate localization of the salient object at each frame. Therefore, it is adaptive to scale and rotation variations, which gives more accurate estimation on both the scale and the position of the target object. 
\setlength{\tabcolsep}{1.8pt}
\begin{table}[tb]
	\renewcommand{\arraystretch}{1.2}
	\caption{Success rates on the 15 sequences of the proposed and the $7$ competing trackers. The best and the second best results are highlighted by bold-face and underline fonts, respectively.}
	\label{Table:SuccessOPE}
	\centering
	\begin{tabular}{lcccccccc}
		\hline
		& Ours &CT &STC &CN &SAMF &DSST &CCT &KCF \\ 
		\hline
		Aircraft &\textbf{0.69} &0.31 &0.18 &0.15 &0.50 &0.15 &\underline{0.55} &0.34 \\   \hline
		airplane\_001 &\textbf{1.00} &0.12 &0.29 &0.14 &\underline{0.36} &0.15 &0.28 &0.02 \\ \hline
		airplane\_006 &\textbf{0.48} &0.34 &0.31 &0.22 &\underline{0.46} &0.37 &\underline{0.46} &0.25 \\ \hline
		airplane\_011 &\underline{0.70} &0.31 &0.30 &0.30 &\textbf{0.99} &0.30 &0.31 &0.30 \\ \hline
		airplane\_016 &0.74 &0.04 &0.74 &\textbf{0.81} &0.66 &0.73 &0.81 &\underline{0.81} \\ \hline
		big\_2 &0.66 &0.38 &0.66 &0.56 &0.56 &\textbf{0.76} &\underline{0.69} &0.26 \\   \hline
		Plane\_ce2 &0.18 &\textbf{0.96} &0.44 &0.14 &\underline{0.47} &0.33 &0.20 &0.31 \\ \hline   
		Skyjumping\_ce &\textbf{0.82} &0.40 &0.18 &0.21 &0.32 &0.13 &\underline{0.40} &0.23 \\ \hline
		motorcycle\_006 &\textbf{0.79} &0.33 &0.21 &\underline{0.40} &0.33 &0.17 &0.17 &0.19 \\ \hline         
		surfing &0.47 &\underline{0.99} &0.99 &0.96 &\textbf{1.00} &\textbf{1.00} &\textbf{1.00} &\textbf{1.00} \\ \hline                  
		Skater &\underline{0.72} &0.28 &0.24 &\textbf{0.73} &0.71 &0.68 &0.61 &0.70 \\ \hline
		Surfer &0.68 &0.31 &0.08 &0.51 &\underline{0.69} &0.65 &\textbf{0.70} &0.25 \\ \hline   	
		Sylvester &0.73 &0.71 &0.71 &0.71 &\textbf{0.97} &0.85 &\underline{0.92} &\textbf{0.97} \\ \hline
		ball &\underline{0.98} &0.72 &0.36 &0.12 &0.97 &0.51 &\textbf{1.00} &0.67 \\ \hline 
		Dog &\textbf{0.93} &0.37 &0.47 &0.35 &0.51 &\underline{0.59} &0.35 &0.38 \\ \hline
		Average success rate &\textbf{0.72} &0.44 &0.41 &0.42 &\underline{0.63} &0.49 &0.58 &0.45 \\ \hline
	\end{tabular}
\end{table}

The second group with target object in high resolution includes three image sequences \textsl{airplane\_011}, \textsl{airplane\_016}, and \textsl{big\_2}, which are illustrated in the fourth to the sixth rows in Fig.~\ref{fig:ImgseqChallenges}. The three image sequences have scale variation and in-plane and out-of-plane rotations challenges. Moreover, there still exists background interference caused by the clouds in the image sequences of \textsl{airplane\_016} and \textsl{big\_2}. The background interference is more severe in the image sequence \textsl{big\_2} as the jet plane is at a lower altitude during its flight. Only the proposed tracker tracks the flying propeller plane accurately with giving the minimum bounding box of the target object in the image sequence of \textsl{airplane\_011}. While most of the competing trackers are able to track the aircraft in the image sequences of \textsl{airplane\_016} and \textsl{big\_2}, only the proposed approach has the capability to give the optimal tracking bounding box in the sky-region scenarios with clouds interference. 

The target object in the third group is a person doing sky jumping, as shown in the seventh row in Fig.~\ref{fig:ImgseqChallenges}. There exists small scale variation and large in-plane and out-of-plane rotations. Moreover, the clouds and terrain of the land in the background would interfere with the tracking performance. The competing trackers STC, SAMF, DSST, and CCT were capable to handle scale changes, but they failed in this image sequence. The competing trackers fail to handle the significant appearance changes of rotating motions and fast scale variations. In contrast, our tracker is robust to large and fast scale variations. 

The proposed tracker can also be used to track the object of interest on the ground with large scale variation and out-of-plane rotation, as shown in the eighth row in Fig.~\ref{fig:ImgseqChallenges}. Since the cycler is driving the motorcycle with large rotation, the competing trackers can only track the head of the motorcycle, however, the proposed tracker is still able to give the minimum tracking bounding box of the object of interest, as shown from frame $\#110$ to frame $\#145$. 

In summary, the proposed tracker has better performance than the seven competing trackers in handling large scale variation, in-plane and out-of-plane rotations with acute angle, and background cloud interference. 

\subsection{Limitation}
\label{sec:limination}
\begin{figure}[tb]
	\centering
	\includegraphics[width=0.48\textwidth]{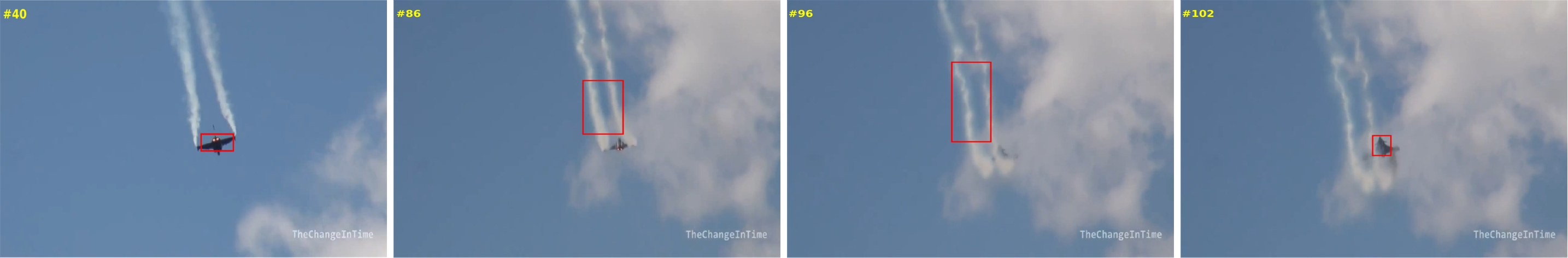} 

	\caption{A failure case where the aircraft is partially or severely occluded in sequence \textsl{airplane\_005}~\cite{li2015nus} by the smoke and cloud.}
	\label{fig:Limitation}
\end{figure}

Although the proposed tracking approach outperforms its competitors in most experiments, it has a key limitation in handling occlusion challenge. As shown in Fig.~\ref{fig:Limitation}, the image sequence \textsl{airplane\_005}, where the aircraft is partially (frame $\#86$) or severely (frame $\#96$) occluded by its ejected smoke and the cloud in the sky, which would cause a tracking failure. This failure can be automatically corrected by the tracker after a few frames, as shown in frame $\#102$. However, this limitation would deteriorate the performance of the tracker. 

\section{Conclusion}
\label{sec:conclusions}
In this paper, we have proposed an effective and efficient approach for real-time visual object localization and tracking, which can be applied to UAV navigation, such as obstacle sense and avoidance. Our method integrates a fast salient object detector within the Kalman filtering framework. Compared to the state-of-the-art trackers, our approach can not only initialize automatically, it also achieves the fastest speed and better performance than the competing trackers. 

Although the proposed tracker performs very well in most image sequences in our experiments, it cannot handle occluded scene very well. However, it has the capability to automatically re-localize and track the salient object of interest when it re-appears in the field of view again. 

\section*{Acknowledgment}
This work is partly supported by the National Aeronautics and Space Administration (NASA) LEARN II program under grant number NNX15AN94N. The authors would like to thank Mr. Arjan Gupta at the University of Kansas for labeling the test data.

\bibliographystyle{IEEEtranS}
\bibliography{reference}

%
%

%

\begin{IEEEbiography}[{\includegraphics[width=1in,height=1.25in,clip,keepaspectratio]{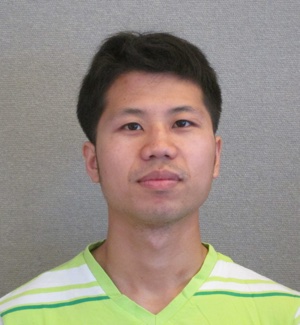}}]{Yuanwei Wu}
	received his Master's degree from the Tufts University. He is currently a PhD candidate at the University of Kansas. His research interests are focused on broad applications in deep learning and computer vision, in particular object detection, localization and visual tracking.
\end{IEEEbiography}
\vspace{-2cm}
\begin{IEEEbiography}[{\includegraphics[width=1in,height=1.25in,clip,keepaspectratio]{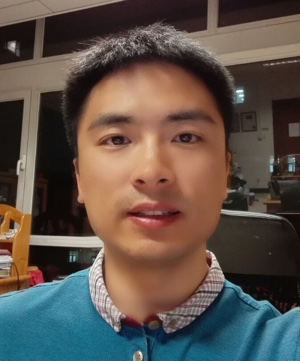}}]{Yao Sui}
	received his Ph.D. degree in electronic engineering from Tsinghua University, Beijing, China, in 2015. He is currently a postdoctoral researcher in the Department of Electrical Engineering and Computer Science, University of Kansas, Lawrence, KS 66045, USA. His research interests include machine learning, computer vision, image processing and pattern recognition.
\end{IEEEbiography}
\vspace{-2cm}
\begin{IEEEbiography}[{\includegraphics[width=1in,height=1.25in,clip,keepaspectratio]{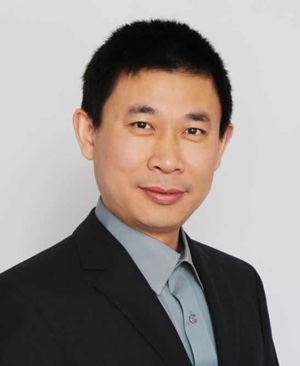}}]{Guanghui Wang (M'10)}
	received his PhD in computer vision from the University of Waterloo, Canada, in 2014. 
	He is currently an assistant professor at the University of Kansas, USA. He is also with the Institute of Automation, Chinese Academy of Sciences, China, as an adjunct professor. 
	
	From 2003 to 2005, he was a research fellow and
	visiting scholar with the Department of Electronic Engineering at
	the Chinese University of Hong Kong. From 2005 to 2007, he acted as a
	professor at the Department of Control Engineering in
	Changchun Aviation University, China. From 2006 to 2010, He was a
	research fellow with the Department of Electrical and Computer
	Engineering, University of Windsor, Canada. He has authored one book
	\emph{Guide to Three Dimensional Structure and Motion
		Factorization}, published at Springer-Verlag. He has published over 80 papers
	in peer-reviewed journals and conferences. His research interests
	include computer vision, structure from motion, object detection and tracking, artificial intelligence, and robot localization and navigation. Dr. Wang has served as associate editor and on the editorial board of two journals, as an area chair or TPC member of 20+ conferences, and as a reviewer of 20+ journals.
\end{IEEEbiography}



\end{document}